\documentclass{article}
\pdfoutput=1

\PassOptionsToPackage{numbers}{natbib}

\usepackage[preprint]{neurips_2018}

\usepackage[]{graphicx}
\usepackage{verbatim}
\usepackage{color}
\usepackage{subfig}
\usepackage[font={small}]{caption}
\usepackage[binary-units]{siunitx}
\usepackage{abbrevs}
\usepackage{pifont}
\usepackage{xspace} 
\usepackage{listings}
\usepackage[normalem]{ulem}
\usepackage{amsmath, amssymb}

\usepackage{hyperref}

\definecolor{red3}{rgb}{0.80,0.00,0.00}

\makeatletter
\let\maybe@space@\xspace
\makeatother

\newcommand{\CM}[1]{}

\newcommand{\snapml}{Snap ML\xspace}

\usepackage{algorithm}
\usepackage{algpseudocode}
\DeclareMathOperator*{\argmin}{arg\,min}

\algblock{ParFor}{EndParFor}
\algnewcommand\algorithmicparfor{\textbf{parfor}}
\algnewcommand\algorithmicpardo{\textbf{do}}
\algnewcommand\algorithmicendparfor{\textbf{end\ parfor}}
\algrenewtext{ParFor}[1]{\algorithmicparfor\ #1\ \algorithmicpardo}
\algrenewtext{EndParFor}{\algorithmicendparfor}

\usepackage{enumitem}

\begin{document}

\title{Parallel training of linear models without compromising convergence}

\author{
Nikolas Ioannou, Celestine D{\"{u}}nner, Kornilios Kourtis, and Thomas Parnell\\
 IBM Research Zurich, Switzerland \\
  \texttt{\{nio,cdu,kou,tpa\}@zurich.ibm.com}
}


\maketitle

\begin{abstract}
In this paper we analyze, evaluate, and improve the performance of training generalized linear models on modern CPUs.
We start with a state-of-the-art asynchronous parallel training algorithm, identify system-level performance bottlenecks, and apply optimizations that improve data parallelism, cache line locality, and cache line prefetching of the algorithm.
These modifications reduce the per-epoch run-time significantly, but take a toll on algorithm convergence in terms of the required number of epochs. To alleviate these shortcomings of our systems-optimized version, we propose a novel, dynamic data partitioning scheme across threads which allows us to approach the convergence of the sequential version.
The combined set of optimizations result in a consistent bottom line speedup in convergence 
of up to $\times12$ compared to the initial asynchronous parallel training algorithm and up to $\times42$, compared to state of the art implementations (scikit-learn and H20) on a range of multi-core CPU architectures.


\end{abstract}

\section{Introduction}




Today's individual machines offer dozens of cores and hundreds of gigabytes of RAM that can, if used efficiently, significantly contribute to improve training performance of machine learning models. In this respect parallel versions of popular machine learning algorithms such a stochastic gradient descent \cite{Recht2011} and stochastic coordinate descent \cite{liu2015asynchronous, Hsieh2015} have been developed. These methods introduce asynchronicity to the sequential algorithms in order to enable parallelization and better utilization of compute resources.  However, these methods treat machines as a simple, uniform, collection of cores. This is far from reality.
While modern machines offer ample computation and memory resources, they are also elaborate systems with complex topologies, memory hierarchies, and CPU pipelines.
As a result, maximizing the performance of parallel training, requires implementations that are aware of these system-level details and address their bottlenecks.

In this paper, we focus on the popular stochastic coordinate descent algorithm \cite{Wright2015, sdca2013} and take a system-aware approach of building a parallel model trainer. We start with a system-oblivious state-of-the-art asynchronous multi-threaded implementation written in OpenMP\cite{snapml18nips}. As a first step, we identify bottlenecks and scalability issues within the execution of a single epoch. We address these issues by modifying the algorithm to be more aligned with the system architecture, leading to a $\times17.3$ faster run-time per epoch on average. However, these modifications come at the cost of convergence; our modifications increase the number of epochs required to converge. To address this, we combine our previous optimizations with a novel dynamic data partitioning algorithm that achieves \emph{both} efficient execution \emph{and} fast convergence. These combined optimizations lead to an average speedup in convergence time of $\times5.1$ compared with our previous implementation, and a speedup of $\times18.1$ on average, when comparing against scikit-learn~\cite{scikit-learn} and H2O~\cite{h2o}.





\begin{algorithm}[b!]
  \begin{algorithmic}[1]
    \small
\State \textbf{Input:} Training data matrix $A=[\mathbf x_1, ... , \mathbf x_n]\in \mathbb{R}^{d\times n }$
	\State Initialize model $\boldsymbol \alpha$ and shared vector ${\mathbf v}=\sum_{i=1}^n \alpha_i {\mathbf x}_i$.
	\For {$t=1,2,\ldots,N_{epochs}$}
		\ParFor {$j\in\Call{RandomPermutation}{n}$}
			\State Read current state of model $\hat{\alpha}_j = \Call{Read}{\alpha_j}$
			\State Read current state of shared vector $\hat{\mathbf v} = \Call{Read}{\mathbf v}$
			\State $\delta = \argmin_{\delta \in\mathbb {R}} f(\hat {\mathbf v} + {\mathbf x}_j \delta)+g_j(\hat \alpha_j+\delta)$
			\State $\Call{Write}{\alpha_j,  \hat{\alpha}_j + \delta}$
			\For {$i=1,2,\ldots,d$}
				\State $\Call{Add}{v_i, \delta A_{i,j}}$
			\EndFor
		\EndParFor
	\EndFor
\end{algorithmic}
\caption{\small SDCA for training GLMs}
\label{alg:a-sdca}
\end{algorithm}

\section{Baseline Implementation}
\label{sec:baseline}
\vspace{-0.1in}
We use the \snapml accelerated ML framework~\cite{snapml18nips} as the basis of this study.
\snapml offers state-of-the art sequential and multi-threaded implementations of SDCA \cite{sdca2013}, with the multi-threaded implementation being an asynchronous parallel training algorithm as detailed in Algorithm \ref{alg:a-sdca}.
The algorithm operates in epochs and repeatedly divides the $n$ shuffled coordinates amongst the parallel threads.  Each thread then operates asynchronously: reading the current state of the model $\mathbf \alpha$, computing an update for this coordinate and writing out the update to the model $\alpha_j$ as well as the shared vector $\mathbf v$.
While no two threads operate on the same coordinate of $\mathbf\alpha$, the shared vector is accessed and updated by all the threads. To avoid an expensive locking mechanism this is done opportunistically in a ``wild'' fashion, i.e., without synchronization.
It was shown in previous studies \cite{Hsieh2015, parnellFGCS18} that this approach performs reasonably well when the probability of concurrent updates to the shared vector is small, e.g., for extremely sparse data-sets, and for small thread counts.
However, as we will see, if the thread count increases, or the solver is deployed in a multi numa-node machine without taking numa affinity into account, the convergence behavior as well as the execution efficiency deteriorates drastically.

\begin{figure*}[t!]
  \subfloat[dense dataset] {
    \includegraphics[width=.50\linewidth]{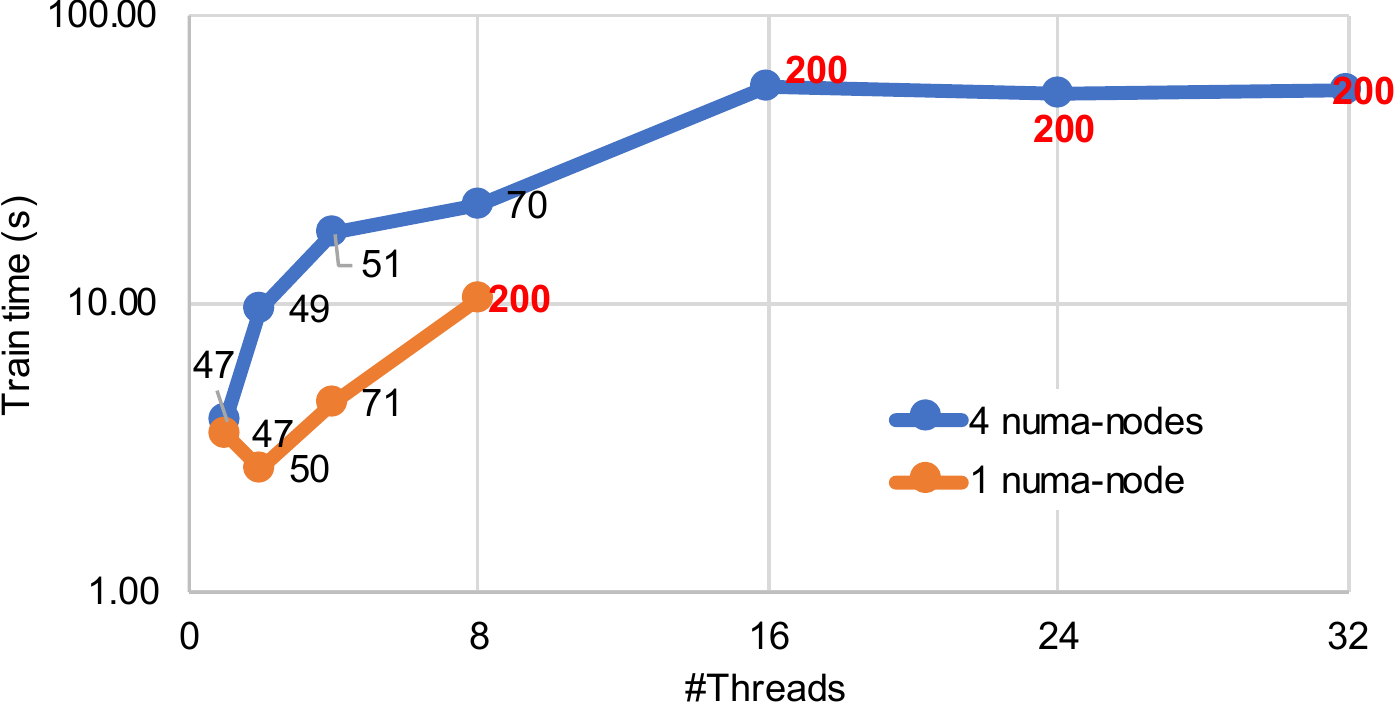}
    \label{fig:mot:dense:x86}
  }
  \subfloat[sparse dataset] {
    \includegraphics[width=.50\linewidth]{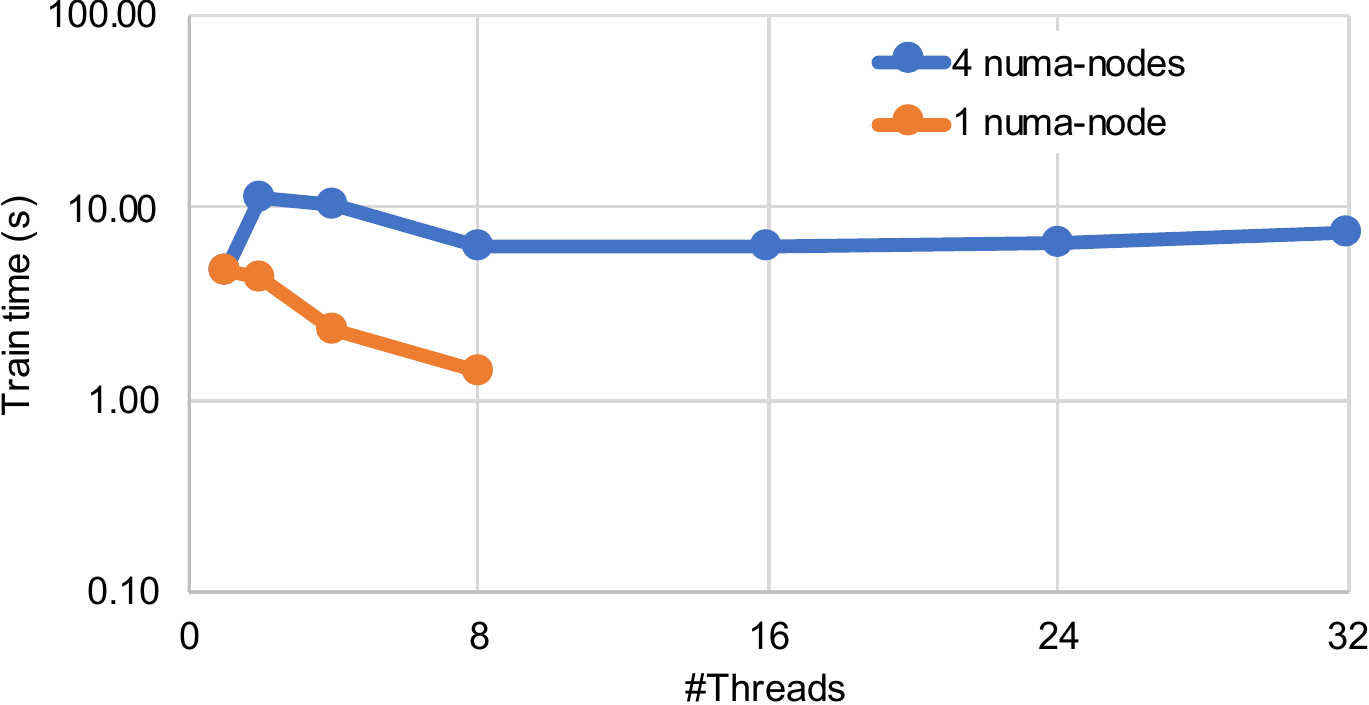}
    \label{fig:mot:sparse:x86}
  }
  \caption{Training time of the ``wild'' multi-threaded SDCA solver on two datasets, running either on one or four numa-nodes. Values in (a) indicate the number of epochs to converge, values in red indicate failure to converge. For the sparse dataset (b) all solvers converged within 15 epochs.}
  \label{fig:mot:example}
  \vspace{-0.15in}
\end{figure*}

To illustrate these issues, we train a logistic regression model on multiple threads using two synthetic datasets of $100$k training examples each: one dense with $100$ features and one sparse with $1$k features and a uniform sparsity of $1\%$.
The results for training these datasets running on one and four numa-nodes are depicted in Fig~\ref{fig:mot:example}.
We see that when running on a single numa node the training on a sparse dataset  (Fig~\ref{fig:mot:sparse:x86}) scales well with the number of threads which is in direct contrast to the training on the dense dataset. This difference can be attributed to highly reduced true sharing among parallel updates to the shared vector for the non-skewed and highly sparse dataset.
When running on multiple numa-nodes, both for the dense (Fig~\ref{fig:mot:dense:x86}) and the sparse (Fig~\ref{fig:mot:sparse:x86}) datasets, we observe a different behavior. The performance of the algorithm is significantly deteriorated because ``wild'' updates on the shared vector result in expensive cache line coherence traffic across the numa nodes. On the dense dataset, this is even more pronounced, due to the higher probability of concurrent updates to the same cache line across the threads.

\section{Optimizing GLM training on CPU}
\label{sec:cpu-opts}
\textbf{Single-Threaded Implementation}
\label{subsec:buckets}
We start by profiling the already vectorized and efficient sequential implementation of the SDCA algorithm.
 Naively, we would expect that for large datasets (e.g., datasets that do not fit in the CPU caches), the run-time would be dominated by a) the inner product computations required for the coordinate update computation and b) retrieving the data from memory, while the ratio between these two would depend on the number of features per training example. 
In our analysis, we have detected two additional bottlenecks:
\vspace{-0.1in}
\begin{enumerate}[leftmargin=0.5cm ]
\setlength\itemsep{0em}
\item When the model does not fit in the cache, a lot of time is spend in accessing the model.
Due to the random nature of the accesses to the model vector, there is very little cache line re-use: a cache line is brought from memory (64B or 128B), out of which only 8B are used.
\item A significant amount of time is spent in the permutation of the example indices before each epoch.
\vspace{-0.05in}
\end{enumerate}

To alleviate these issues, we introduce the concept of \textit{buckets}.
We partition the training examples $\mathbf x_i$ into buckets, and train a bucket of consecutive training examples at a time.
The bucket size is chosen at run-time based on the cache line size of the CPU, using linux \texttt{sysfs}. 
This modification to the algorithm improves performance in several respects; (i) the model vector $\mathbf \alpha$ is accessed in a cache-line efficient manner, (ii)  the amount of indices to be randomized is reduced by a factor equal to the bucket size (e.g., 8- or 16-fold), and (iii)  CPU prefetching efficiency on accessing the different coordinates of the training examples is implicitly improved.

Note that the bucket can decrease the randomness with which training examples are chosen, and thus degrade the convergence of the algorithm. This is especially true for datasets with a small number of training examples. However, as we will see in the experimental section, this trade-off pays off. 
Further, we observe that the bucket optimization reaps limited benefits if the model vector is small enough to fit in the last level cache of the CPU. We thus dynamically choose the bucket size at run-time: if the model vector does not fit in the last level cache of the CPU (typically this cut-off point is in the range of $500$k entries), we use the buckets, otherwise we don't.

\begin{figure*}[t!]
  \subfloat[Multi-threaded performance bottlenecks] {
    \includegraphics[width=.50\linewidth]{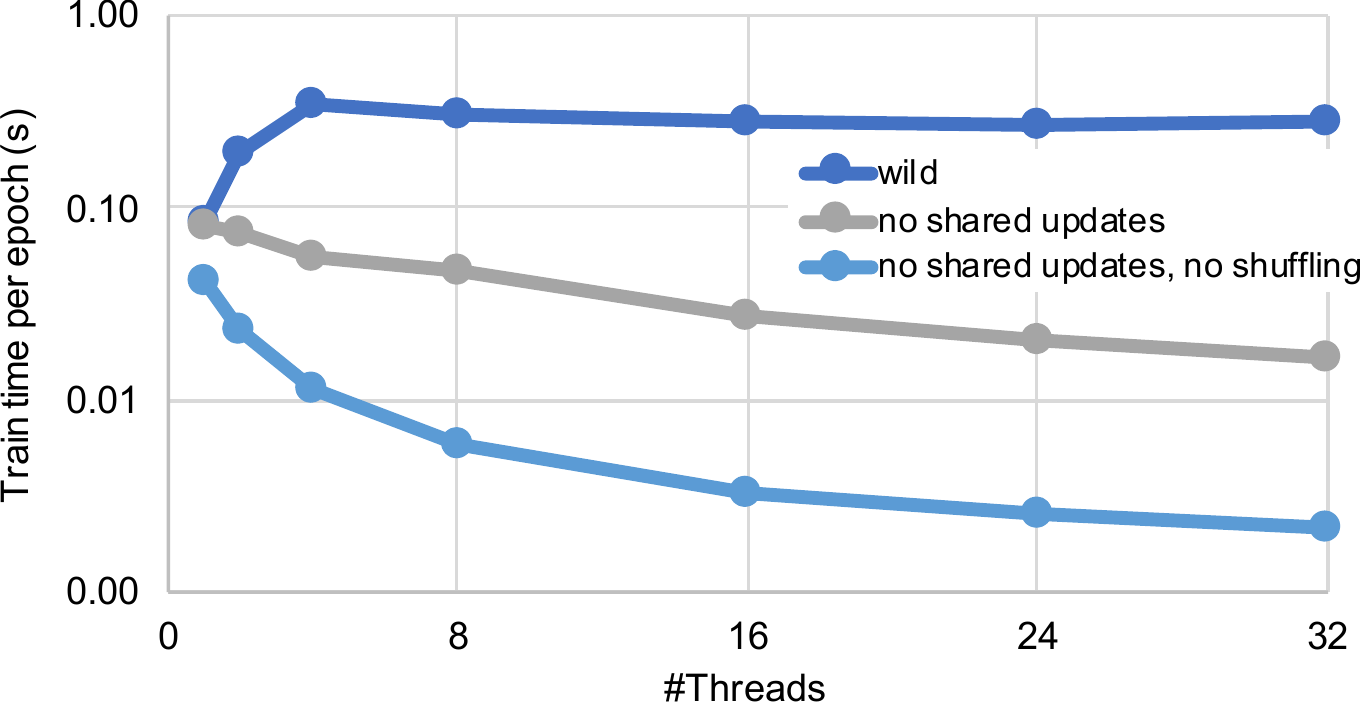}
    \label{fig:mot:dense:scaling:x86}
  }
  \subfloat[Convergence in epochs] {
    \includegraphics[width=.50\linewidth]{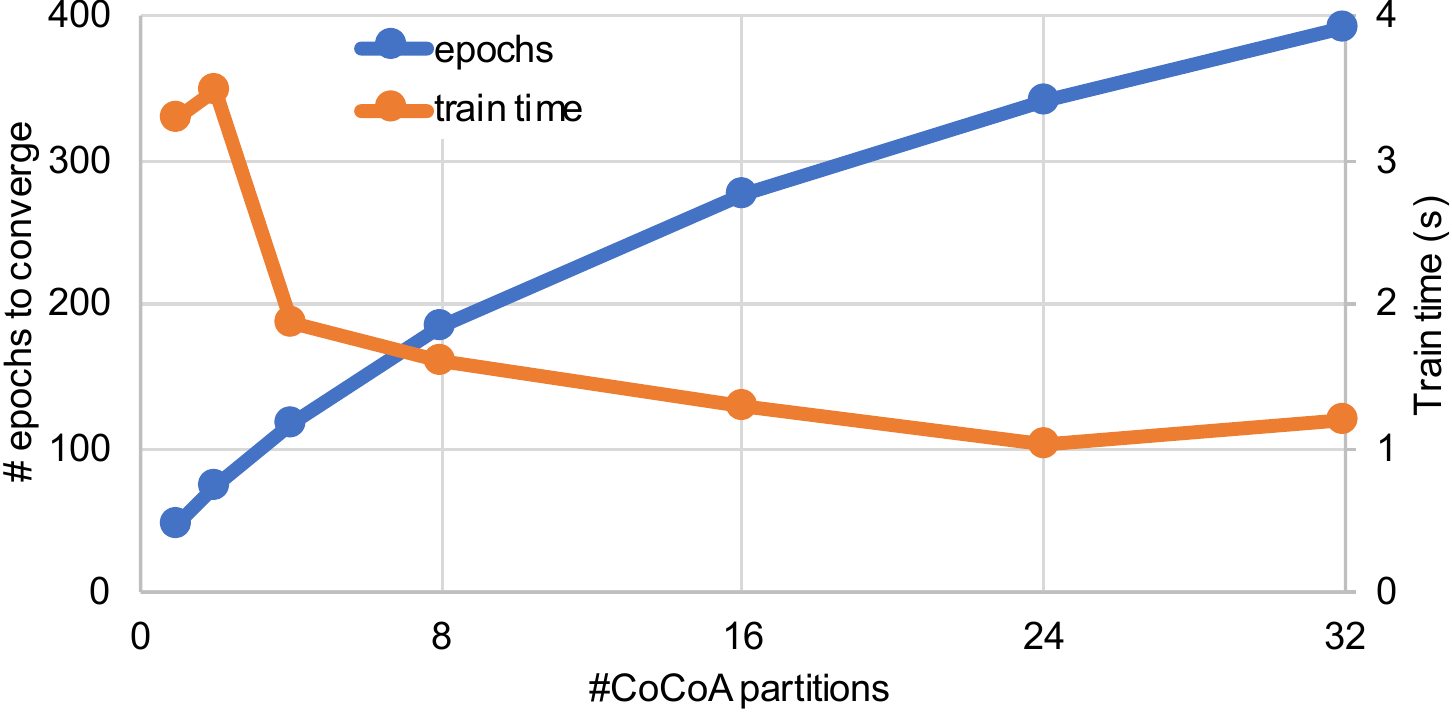}
    \label{fig:mot:cocoa:convergence:x86}
  }
    \caption{(a) Multi-threaded performance of the original algorithm without shared updates, and without shuffling, on the dense artificial dataset. (b) Effect of increasing the number of CoCoA partitions (1 partition per thread) in number of epochs and time to converge for the same dataset.}
  \label{fig:mot:example:2}
\vspace{-0.2in}
\end{figure*}

\textbf{Multi-threaded Implementation}
\label{subsec:data-par}
We now turn to a asynchronous multi-threaded implementation of our optimized sequential SDCA algorithm.
Unsurprisingly, the shared vector updates over shared memory across the threads turns out to be a scalability bottleneck: simply disabling those updates results in improved scaling, as depicted in Fig~\ref{fig:mot:dense:scaling:x86}.
The next scalability bottleneck is the sequential shuffling of the training example indices, also shown in Fig~\ref{fig:mot:dense:scaling:x86}.

Based on these observations, we propose to increase data parallelism of the algorithm to improve scalability.
To achieve this we transfer ideas from distributed learning \cite{cocoa18jmlr}, where the training examples are partitioned across worker nodes that independently work on a local version of the shared vector which is synchronized periodically.
We map this approach to a parallel architecture where we partition the examples across the threads and replicate the shared vector in each one.
In this way we have achieved that the global shared vector needs to be accessed by different  threads much less frequently.

Again, this system-optimization has a price -- convergence suffers. The static partitioning of the training examples across processes is known to increase the epochs needed to convergence \cite{cocoa18jmlr}. We have illustrated this at a toy example in Figure \ref{fig:mot:cocoa:convergence:x86}. In order to alleviate this issue, we propose a \textit{dynamic partitioning} for the multi-threaded implementation: we shuffle all the (buckets of) examples at the beginning of each epoch, and each thread picks a different set of buckets in each epoch. Such a re-partitioning approach is very effective but has not been adapted by distributed algorithms previously because it is too expensive in a distributed environment.

\textbf{Numa-level optimizations}
\label{subsec:numa-opt}
Subsequently, we focus on optimizations related to numa topology in a multi numa node system.
We treat each node as an independent training node in a distributed setting, and deploy a hierarchical scheme: statically partition the training examples across the nodes in a distributed fashion, and within the numa nodes perform the dynamic partitioning introduced in Sec~\ref{subsec:data-par}.
We exploit the fact that the training dataset is read-only and thus it does not incur expensive coherence traffic across nodes and do not replicate the training dataset across the nodes.
Each node holds its own replica of the shared vector, which is reduced across nodes at the end of each epoch.
The model vector is also local to each node which holds the coordinates corresponding to the part of the training examples it handles. 

We dynamically detect the numa topology of the system, as well as the number of physical cores per node, using \texttt{libnuma} and the \texttt{sysfs} interface.
If the number of threads requested by the user is less or equal to the number of cores in one node, we schedule a single node solver. Otherwise, we evenly distribute the requested number of threads to the minimum number of nodes that can accommodate them w.r.t. physical cores.
We detect the numa node on which the dataset resides using the \texttt{move\_pages} system call, and always include that node in our selection.


\section{Evaluation}

In this section, we evaluate the performance of our optimized implementation within the Snap ML framework in single-server multi numa-node environments.
First, we compare the different multi-threaded implementations and investigate the performance sensitivity of the implementation to the optimizations introduced in Sec~\ref{sec:cpu-opts}.
And then we compare with the widely-used scikit-learn~\cite{scikit-learn} ML framework (0.19.2), as well as the H2O framework~\cite{h2o} (3.20.0.8).
\footnote{We also tried to evaluate VowpalWabbit~\cite{vowpal-wabbit} using their scikit-learn API, but were unable to do so primarily due to the data conversion in the \texttt{tovw} function, which is part of the \texttt{fit} function and takes a significant amount of time (e.g., 4800s for higgs), altering the training performance time.}

We use two systems with different CPU architectures and numa topologies: a 4-node Intel Xeon (E5-4620) with 128GiB of RAM at each node, 512GiB total, and a 2-node IBM POWER9 with 512GiB at each node, 1TiB total.
We disable simultaneous multi-threading
and fix the CPU frequency to the maximum supported (2.2GHz for x86, and 3.8GHz for P9).
We evaluate against 3 datasets: (i) the sparse dataset released by Criteo Labs as part of their 2014 Kaggle competition \cite{criteo} (criteo-kaggle), (ii) the dense HIGGS dataset~\cite{higgs14nature} (higgs), and (iii) the dense epsilon dataset from the PASCAL Large Scale Learning Challenge \cite{epsilondataset} (epsilon).
Data loading time is not included in the training time in any of the results.

\begin{figure*}[t]
  \subfloat[criteo-kaggle] {
    \includegraphics[width=.33\linewidth]{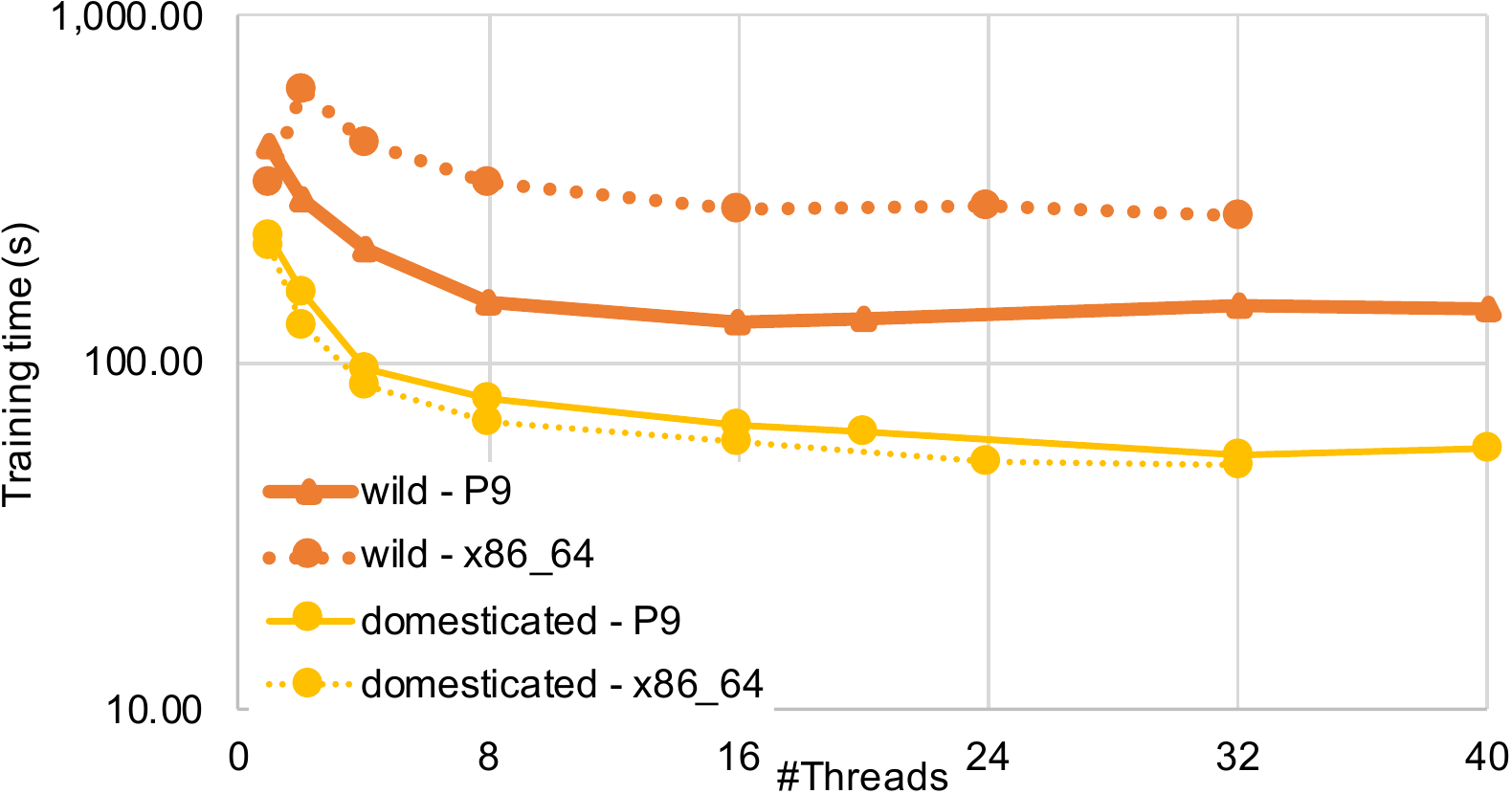}
    \label{fig:tconv:criteo:x86}
  }
  \subfloat[higgs] {
    \includegraphics[width=.33\linewidth]{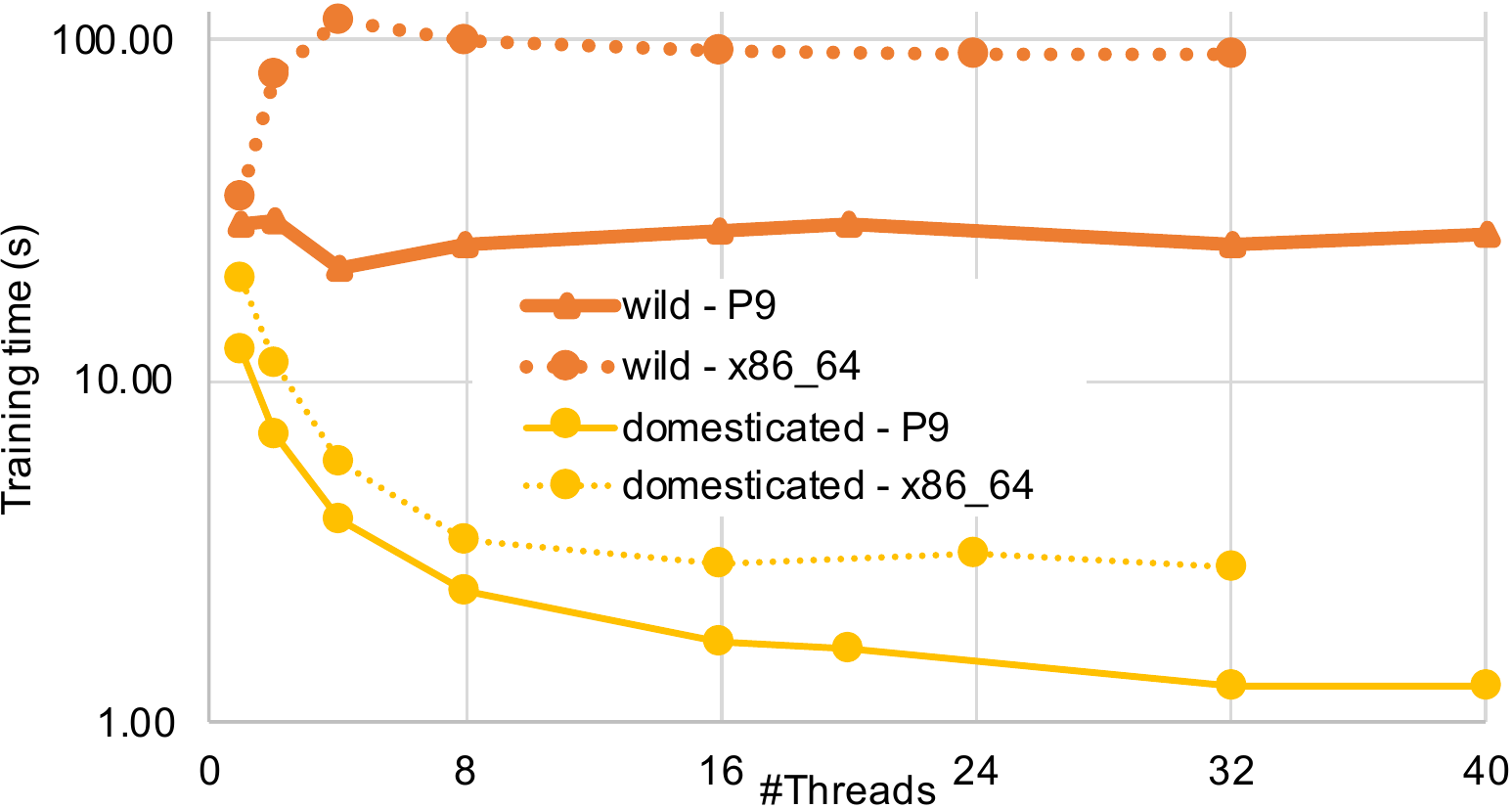}
    \label{fig:tconv:higgs:x86}
  }
  \subfloat[epsilon] {
    \includegraphics[width=.33\linewidth]{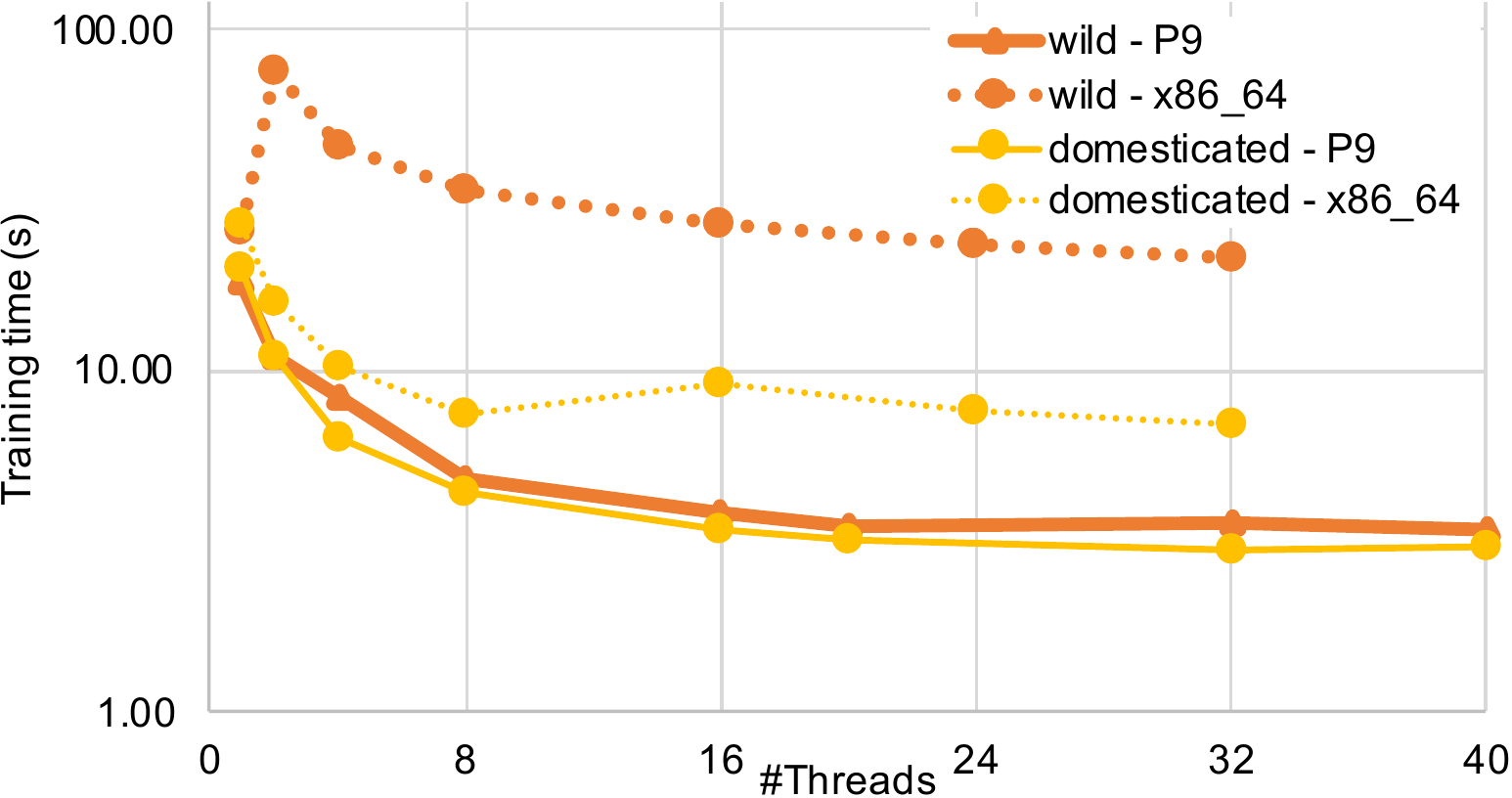}
    \label{fig:tconv:epsilon:x86}
  }

  \caption{Time to convergence as a function of the thread count for the different CPU implementations across different datasets on the two machines.}
  \label{fig:tconv}
\vspace{-0.15in}
\end{figure*}

\textbf{Bottom line performance.}
\label{eval:time-per-epoch}
First, we evaluate the performance in terms of time to convergence between the ``wild'' implementation and our new ``domesticated'' one.  Convergence is declared if the relative change in the learned model form one epoch to the next is below a threshold. We have verified that all implementations exhibit the same test loss after training, apart from the ``wild'' implementation which can converge to an incorrect solution when using many threads~\cite{passcode}.
Fig~\ref{fig:tconv} illustrates the results for all 3 datasets, across the two different systems.
Comparing against the best ``wild'' version that converges to a similar test loss (4 and 8 threads, for the 4 and 2 node systems), the ``domesticated'' optimizations result in a speedup of $\times4.1$, $\times6.8$, and $\times3.4$, for the criteo-kaggle, higgs, and epsilon, respectively.
For the 2 node machine, the speedups are $\times2.3$, $\times11.8$, and $\times1.9$, for the criteo-kaggle, higgs, and epsilon, and $\times5.1$ on average.
The ``wild'' implementation exhibits significantly better performance on the 2 node system relative to the 4 node: this is due to increased memory bandwidth.

\begin{figure*}[t]
  \subfloat[x86\_64] {
    \includegraphics[width=.50\linewidth]{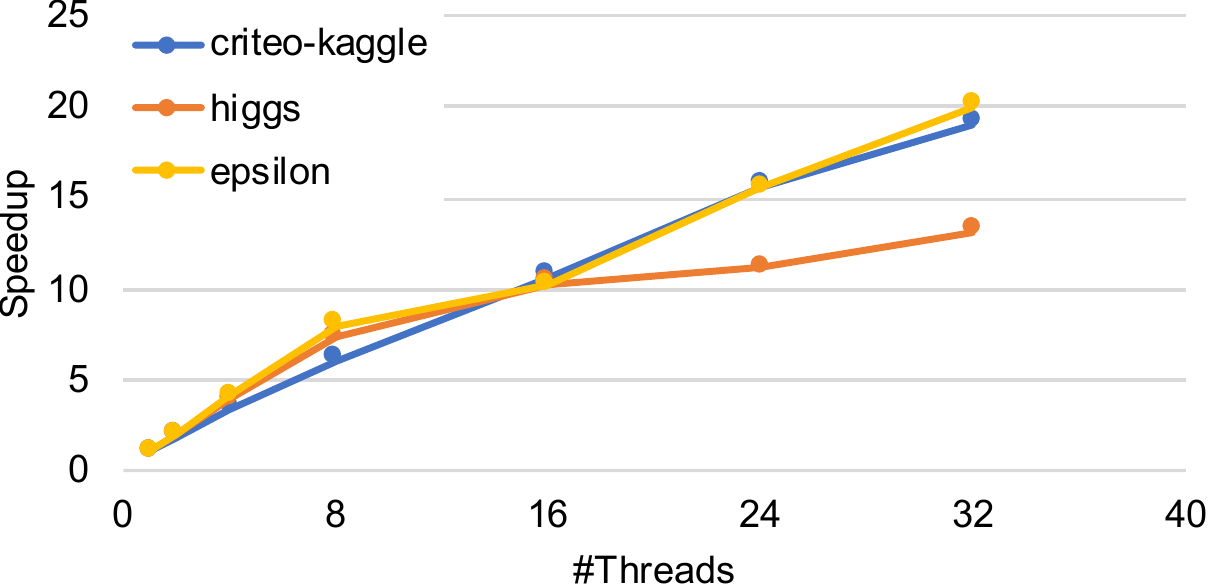}
    \label{fig:tepoch:scalability:x86}
  }
  \subfloat[P9] {
    \includegraphics[width=.50\linewidth]{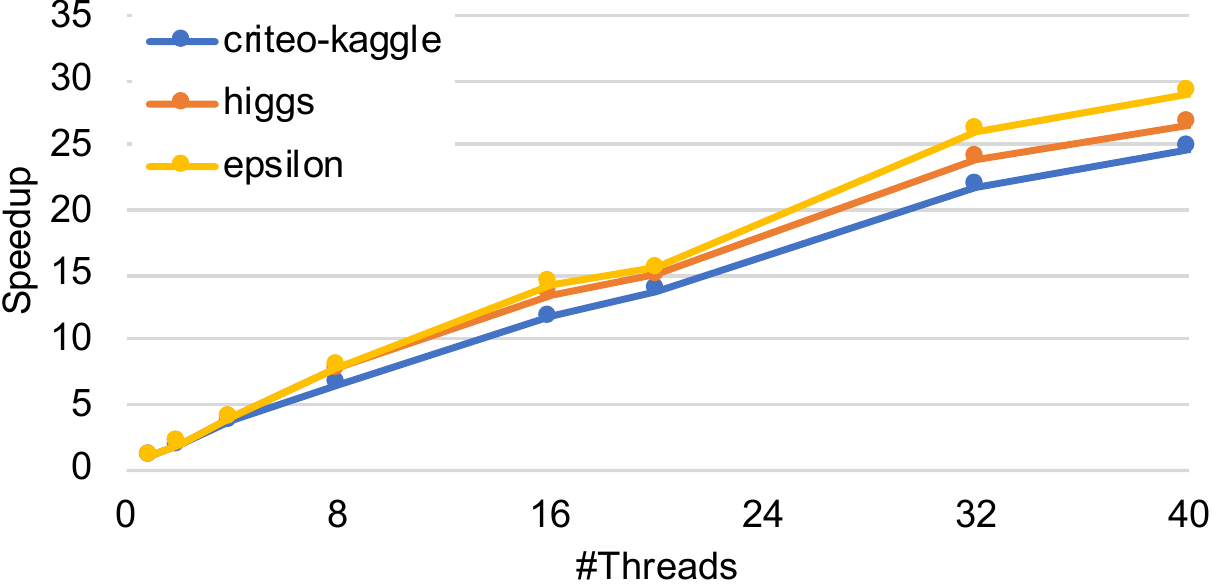}
    \label{fig:tepoch:scalability:P9}
  }
  \caption{Strong scalability w.r.t training time per epoch with increasing thread counts.}
  \label{fig:tepoch:scalability}
\vspace{-0.2in}
\end{figure*}

\textbf{Scalability.}
\label{eval:time-to-acc}
Second, we focus on the strong scalability behavior of the ``domesticated'' implementation w.r.t. time per epoch.
Results, showing the speedup over the sequential version, are depicted in Fig~\ref{fig:tepoch:scalability}.
Performance scales almost linearly for all the datasets, across the two systems.
The 4 node system show a slightly lower absolute speedup beyond 1-node (8 threads), which is expected due to the higher overhead when accessing memory on different numa nodes compared to the 2 node system.
Training on Higgs on the 4 node machine is an exception: scaling gradually degrades when going from 1 to 2 and more numa nodes. By profiling its run-time, we observe that most of the time is spent on memory accesses to the training dataset. 
On the 2 node system, however, which has higher memory bandwidth and less numa nodes, those memory accesses are no longer the bottleneck.

\begin{figure*}[t]
  \subfloat[Static and dynamic partitioning] {
    \includegraphics[width=.33\linewidth]{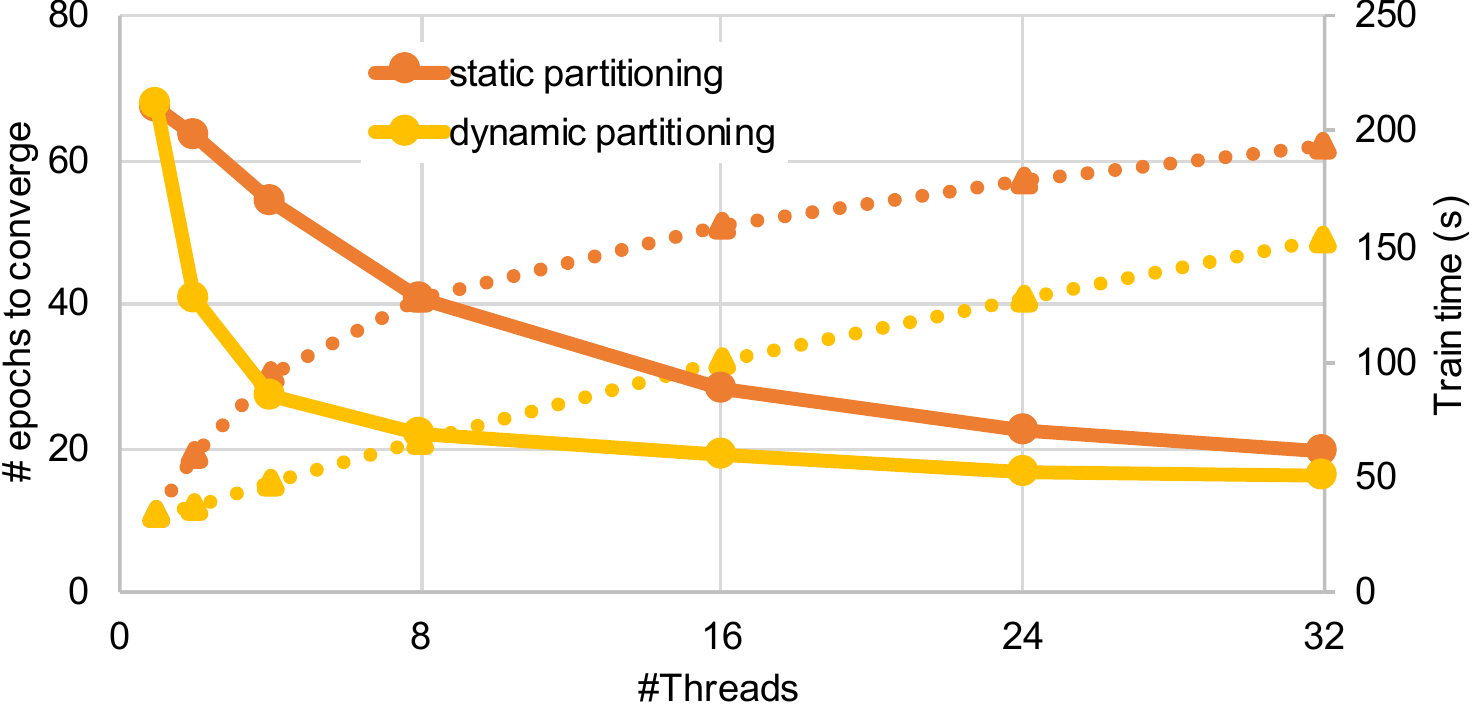}
    \label{fig:eval:shuffle:criteo:x86}
  }
  \subfloat[Bucket optimization] {
    \includegraphics[width=.33\linewidth]{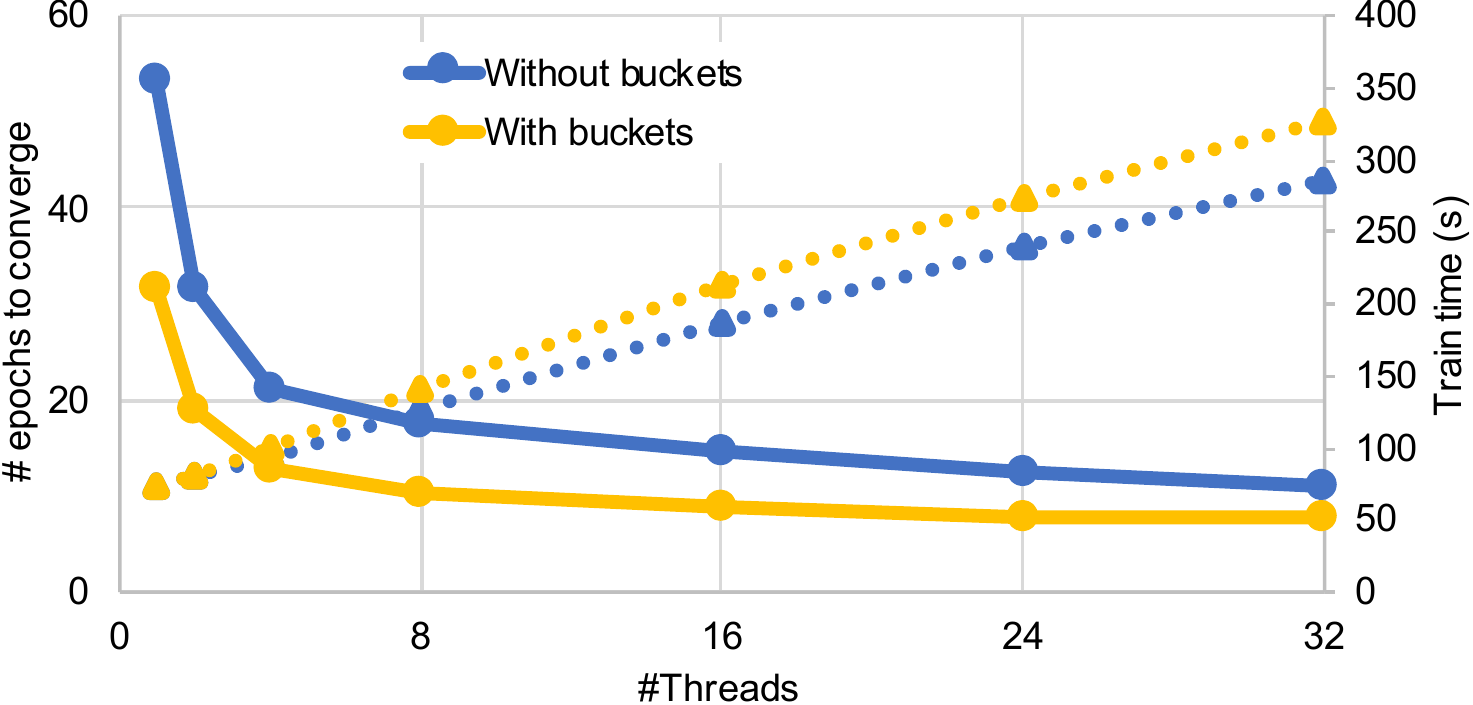}
    \label{fig:eval:bucketsize:criteo:x86}
  }
  \subfloat[Numa optimizations] {
    \includegraphics[width=.33\linewidth]{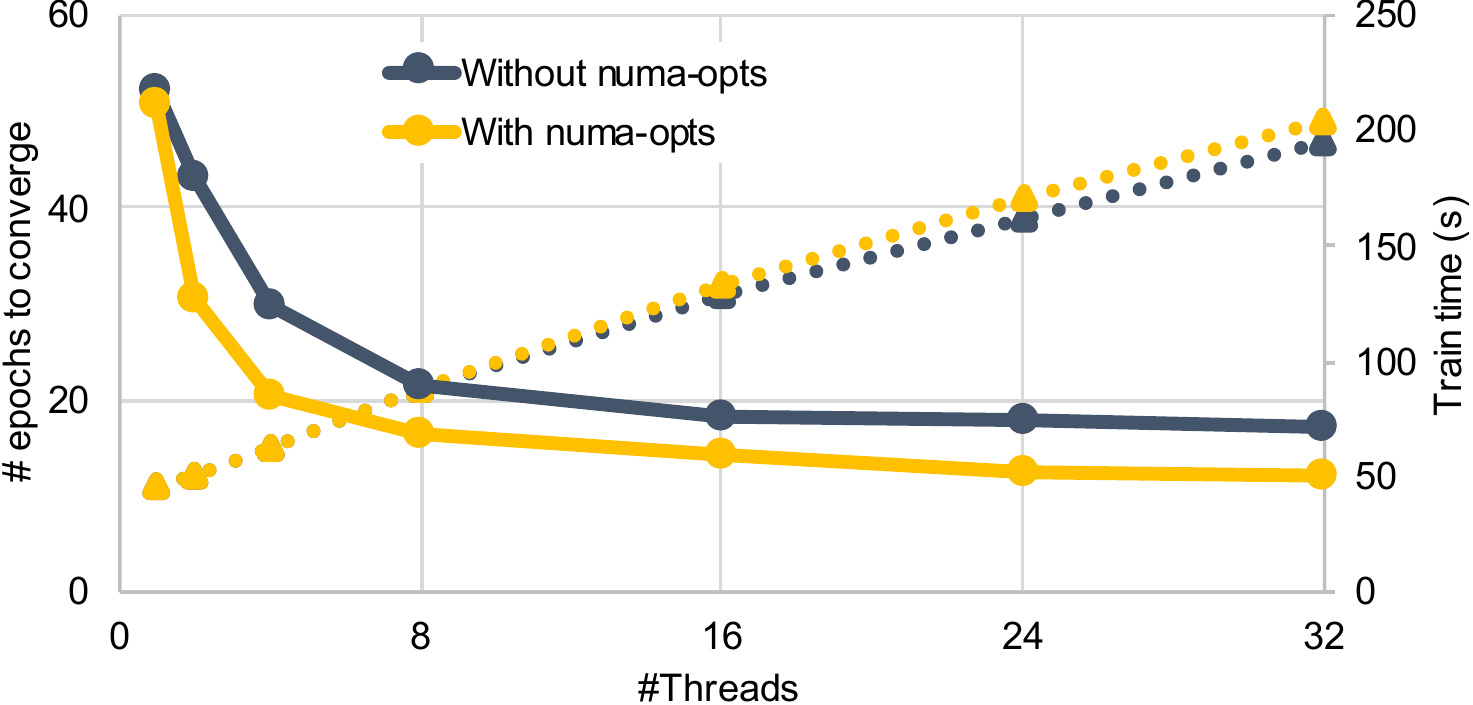}
    \label{fig:eval:numa:criteo:x86}
  }
  \vspace{-0.05in}
    \caption{Evaluation of the gains achieved with our proposed optimizations on the criteo-kaggle dataset on the 4 node system: (a) static vs. dynamic partitioning, (b) bucket optimization, and (c) numa-level optimizations. Solid lines indicate time, and dashed-lines depict number of epochs.}
  \label{fig:eval:opts}
\vspace{-0.15in}
\end{figure*}

\textbf{Training example data partitioning.}
\label{eval:shuffle}
We now evaluate the effect of the dynamic data partitioning scheme, presented in Sec~\ref{subsec:data-par}, against a default static partitioning.
Fig~\ref{fig:eval:shuffle:criteo:x86} compares the two schemes on the criteo-kaggle dataset, for the 4 node system.
By dynamically shuffling the training examples across worker threads within each node after every epoch we are able to gain an improvement in total training of $49\%$ on average compared to the static partitioning, realizing most of the achieved $54\%$ average reduction in epochs.
A similar improvement holds for the epsilon dataset, with an average improvement of $67\%$.
Higgs, however, is less sensitive to data partitioning choices, with the different schemes performing virtually the same.
Similar observations are made for the 2 node machine ($37\%$ and $59\%$, for criteo-kaggle and epsilon).

\textbf{Buckets.}
\label{eval:bucketsize}
Next, the bucket optimization is evaluated in Fig~\ref{fig:eval:bucketsize:criteo:x86}.
This optimization results in an average speedup of $63\%$ for criteo-kaggle 
and $69\%$ for higgs.
When training on epsilon we can not benefit from this optimization at all, since it only has $~300$k training examples, and the model vector completely fits in the last-level cache of the CPU.
Based on the heuristic described in Sec~\ref{subsec:buckets}, our implementation doesn't apply this optimization to epsilon.
The 2 node machine exhibits similar results: $40\%$ and $89\%$ speedup, for criteo-kaggle and higgs.

\textbf{Numa optimizations.}
\label{eval:bucketsize}
We now focus on the impact of the numa optimizations on performance.
Results depicted in Fig~\ref{fig:eval:numa:criteo:x86}, indicate a speedup of $33\%$ on criteo-kaggle.
The numa-level optimizations result in an improvement of $11\%$ and $21\%$, on average, for higgs and epsilon, on the 4 node machine.
These optimizations had a smaller impact on the 2 node machine, with an average speedups of $22\%$, $4\%$, and $2\%$, for criteo, higgs, and epsilon, respectively.

\textbf{Comparison with scikit-learn and H2O.}
\label{eval:sklearn}
Last, we compare the performance of our solver for training a logistic regression model against the widely used scikit-learn~\cite{scikit-learn} library, implementing different solvers (\texttt{liblinear}, \texttt{lbfgs}, \texttt{sag}).
Also, we compare with H2O~\cite{h2o}, using its multi-threaded \texttt{auto} solver
\footnote{We could not get the binary files for the sparse dataset (criteo-kaggle) to work with H2O in a reasonable amount of time and leave this for future work.}.

\begin{figure*}[t]
  \subfloat[criteo-kaggle - x86\_64] {
    \includegraphics[width=.33\linewidth]{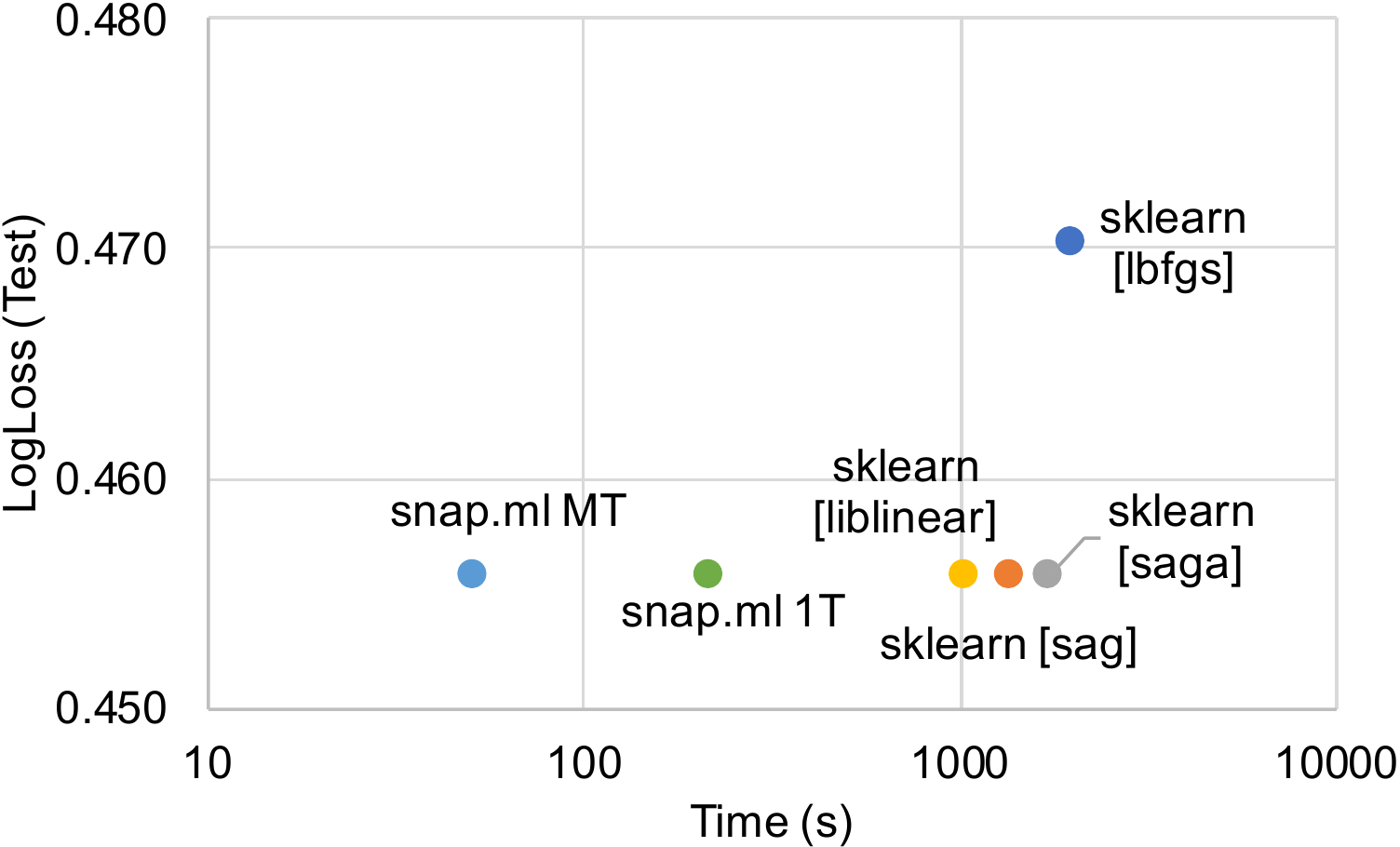}
    \label{fig:ttacc:criteo:x86}
  }
  \subfloat[higgs - x86\_64] {
    \includegraphics[width=.33\linewidth]{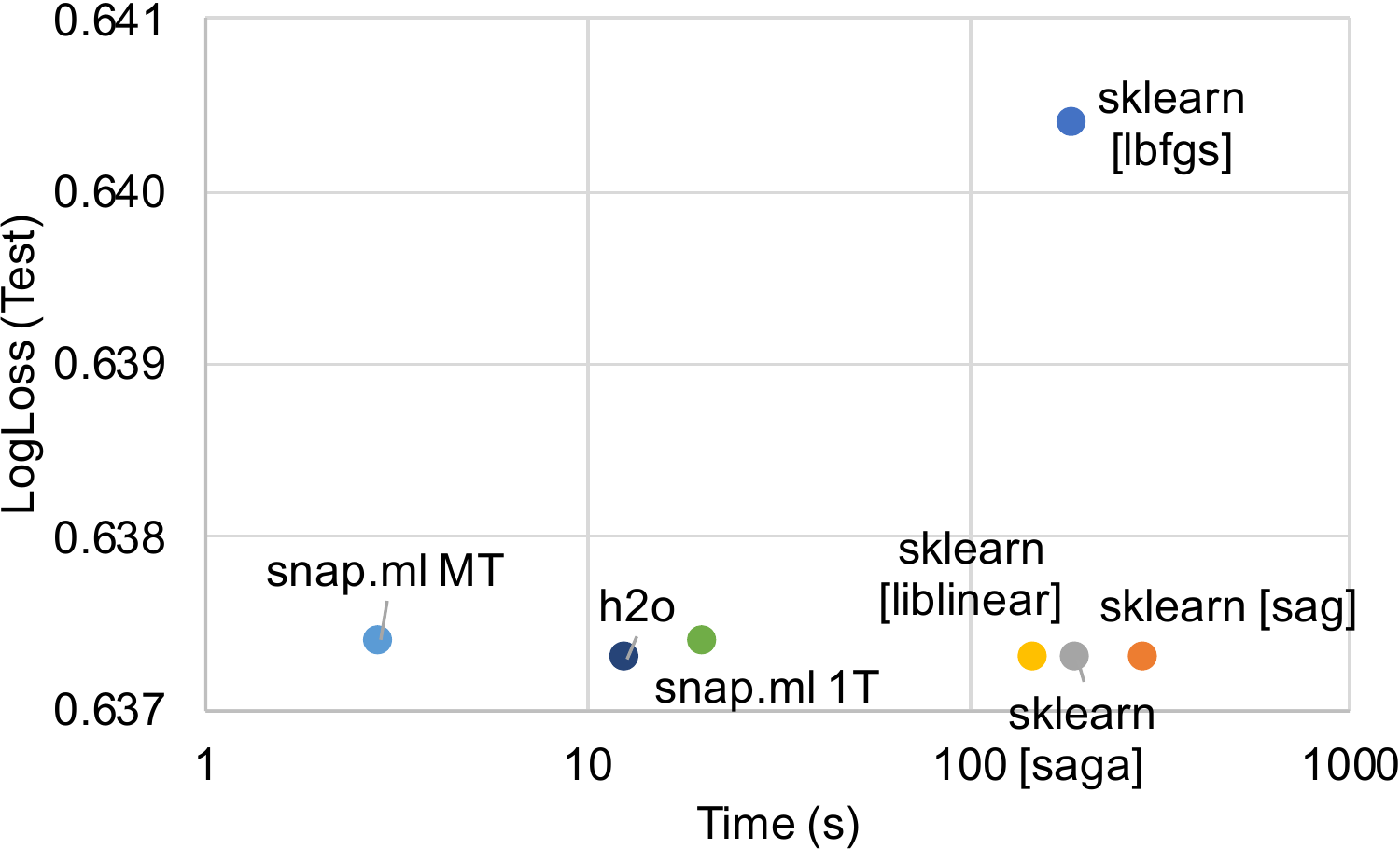}
    \label{fig:ttacc:higgs:x86}
  }
  \subfloat[epsilon - x86\_64] {
    \includegraphics[width=.33\linewidth]{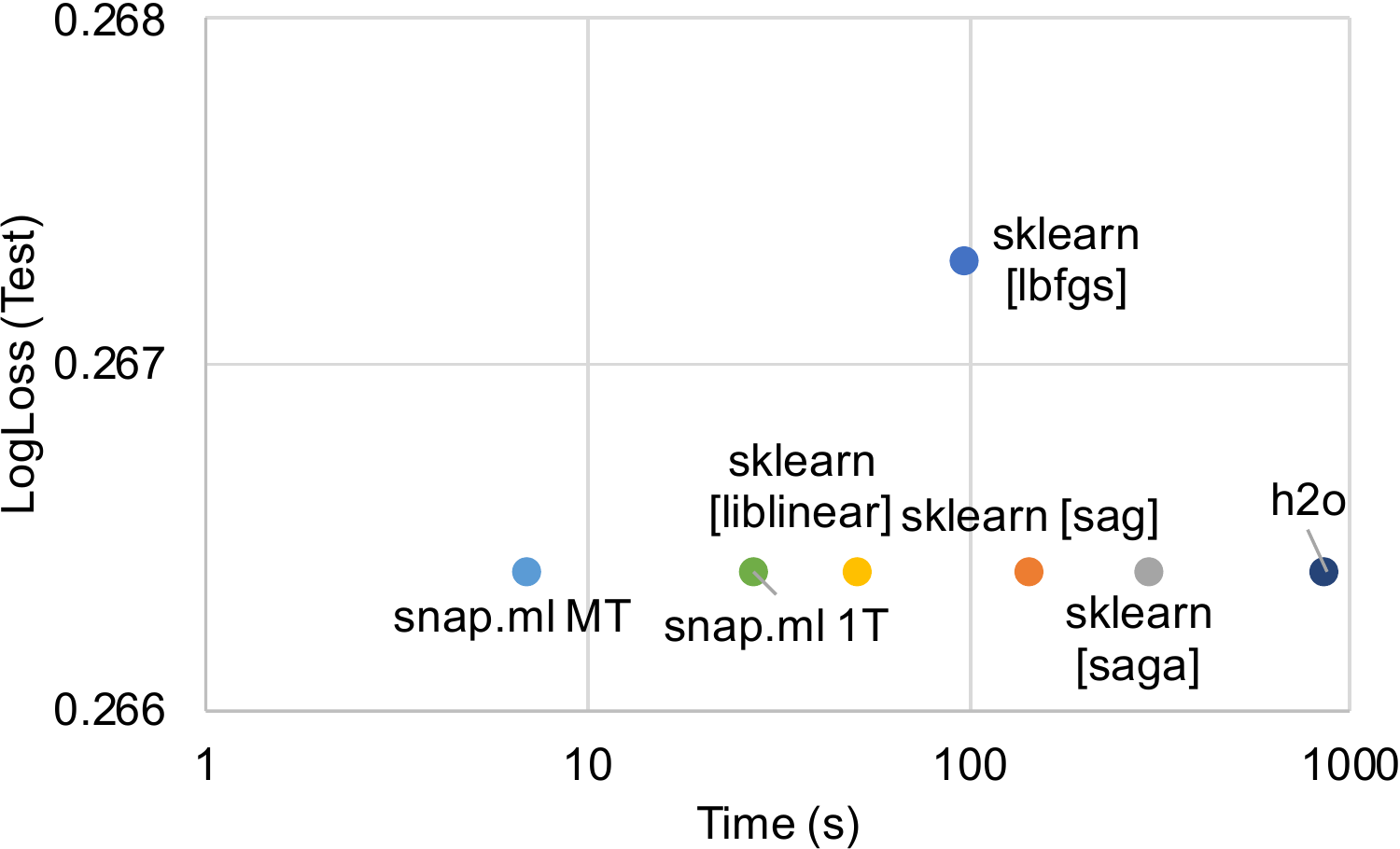}
    \label{fig:ttacc:epsilon:x86}
  }
  \vspace{-0.05in}
\\
  \subfloat[criteo-kaggle - P9] {
    \includegraphics[width=.33\linewidth]{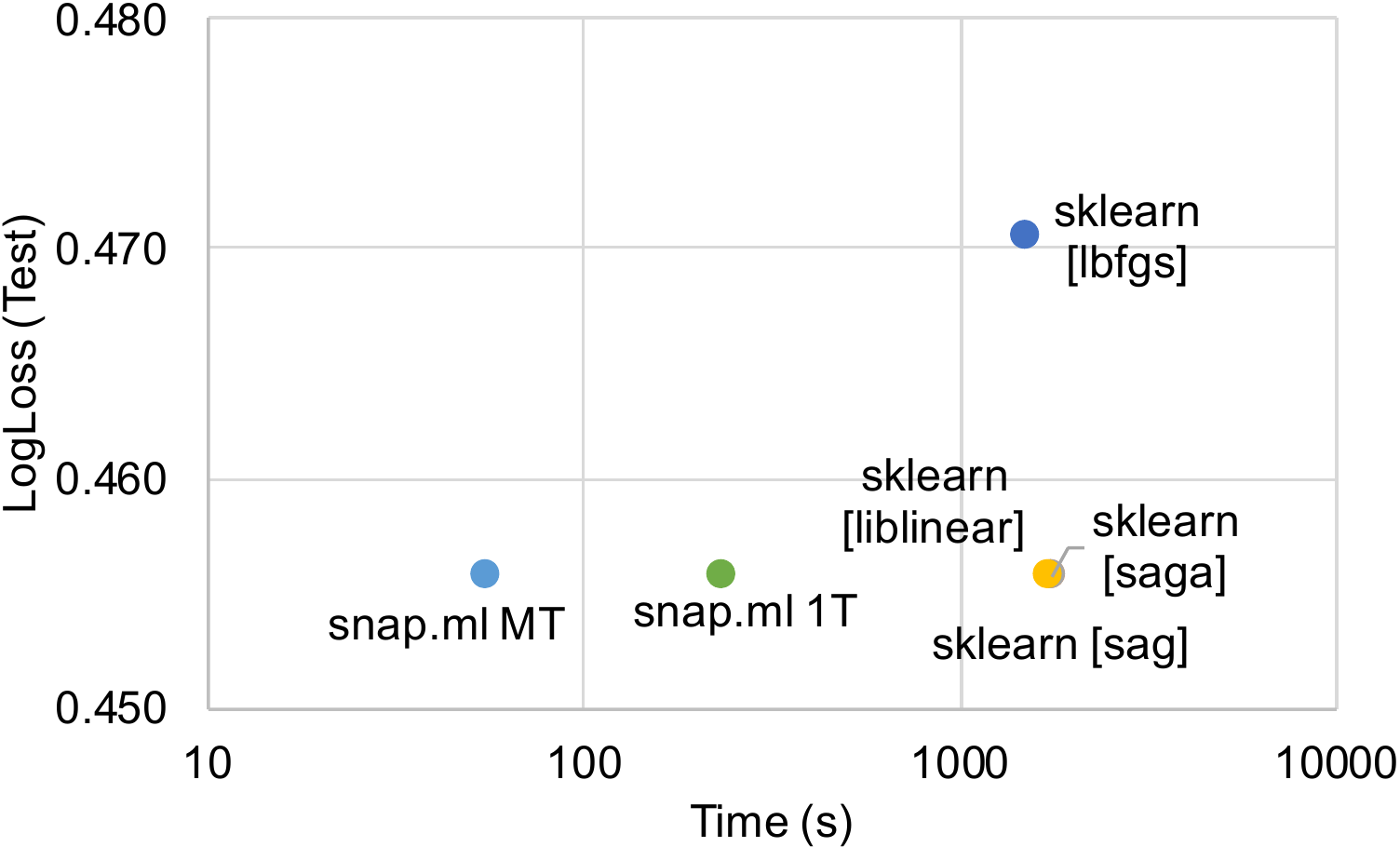}
    \label{fig:ttacc:criteo:P9}
  }
  \subfloat[higgs - P9] {
    \includegraphics[width=.33\linewidth]{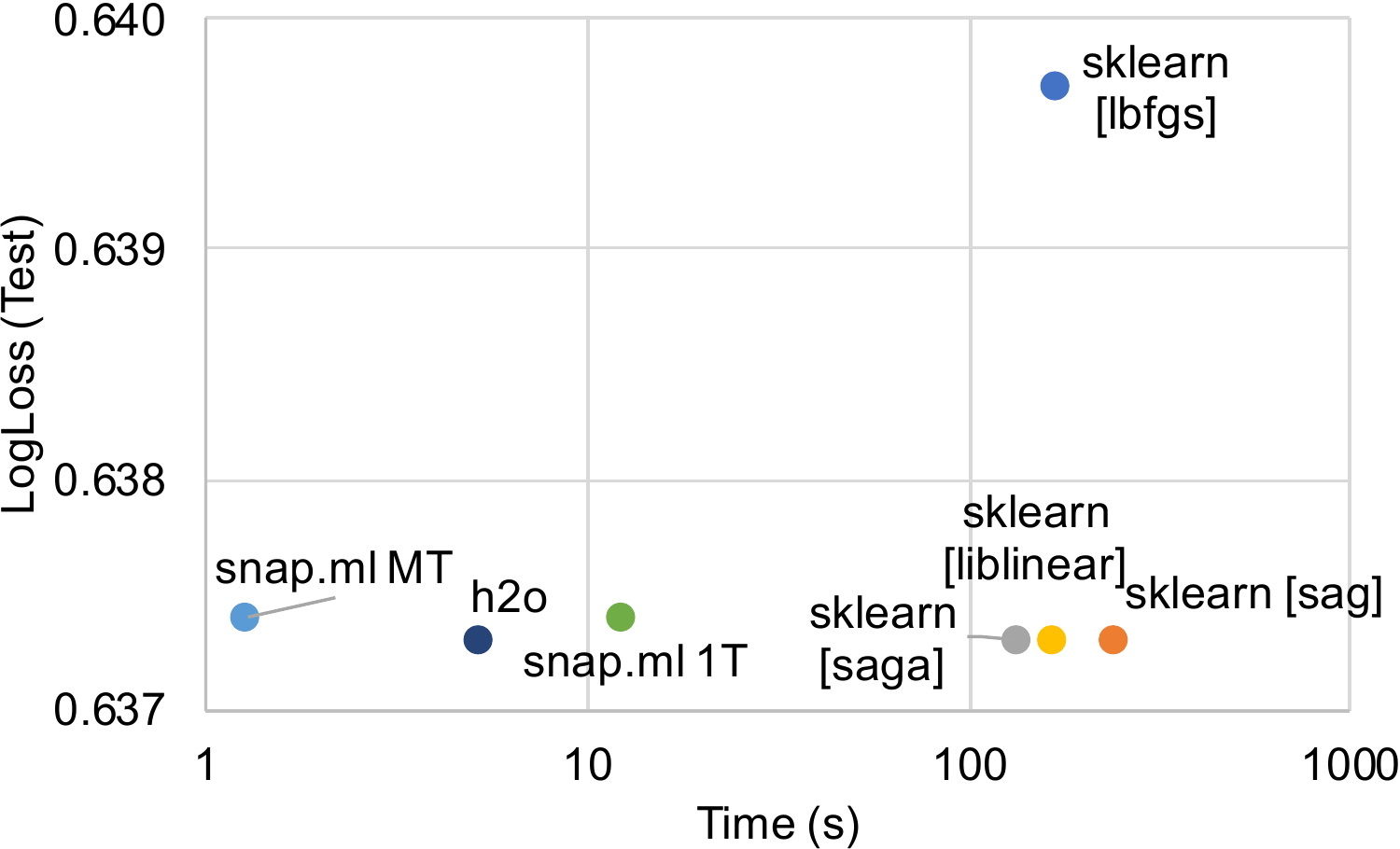}
    \label{fig:ttacc:higgs:P9}
  }
  \subfloat[epsilon - P9] {
    \includegraphics[width=.33\linewidth]{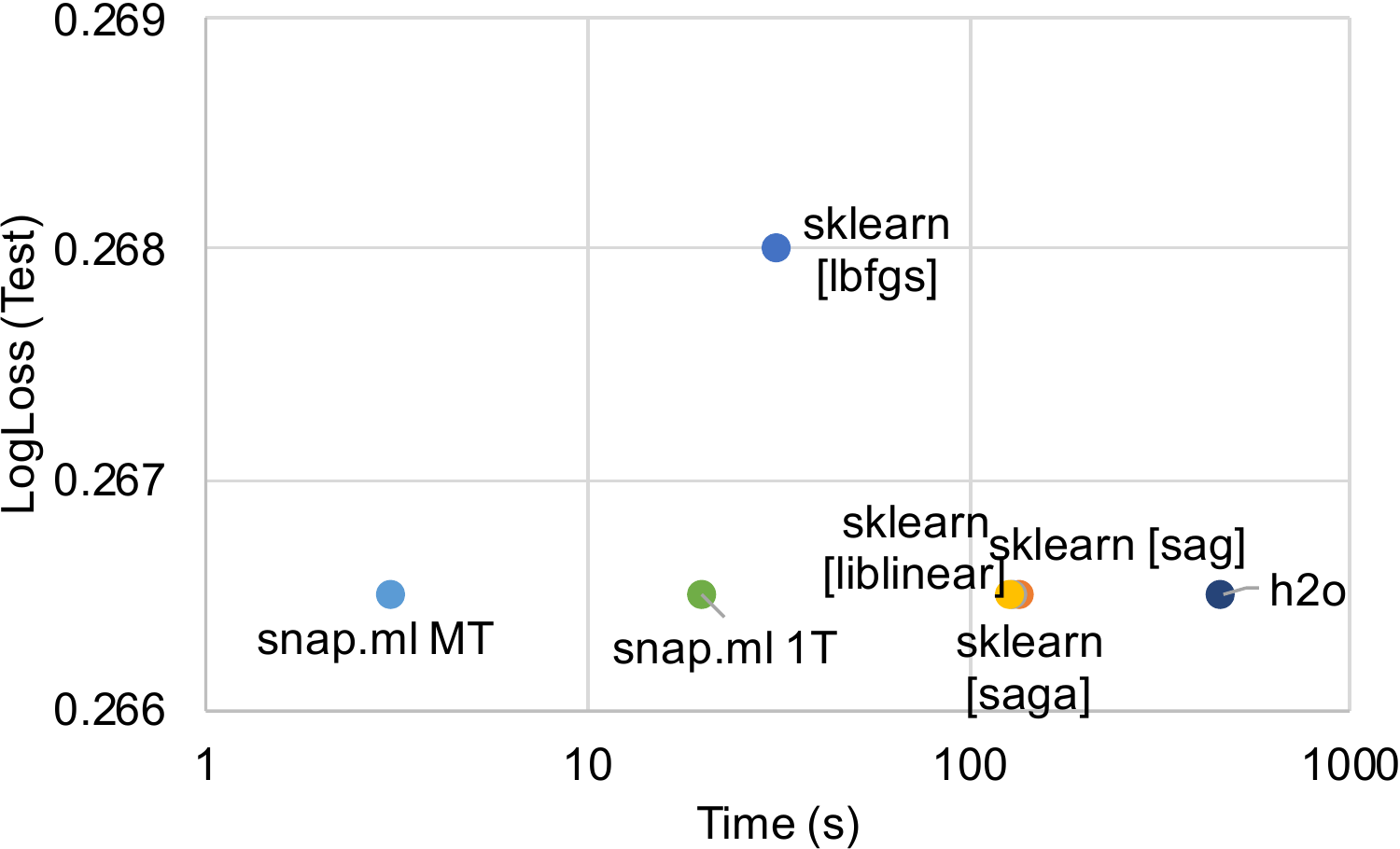}
    \label{fig:ttacc:epsilon:P9}
  }
  \vspace{-0.05in}
  \caption{Comparing our single- and multi-threaded implementations against different solvers in scikit-learn.}
  \label{fig:ttacc}
\vspace{-0.2in}
\end{figure*}

Results comparing the training time against test loss for the different solvers, on the two systems, are depicted in Fig~\ref{fig:ttacc}.
We use results for single (\textit{snap.ml 1T}) and maximum (\textit{snap.ml MT}) thread counts for our optimized implementation.
\textit{snap.ml MT} is consistently faster than the best performing alternative solver, across the board: $\times30.7$, $\times4.1$, and $\times41.7$ for criteo-kaggle, higgs, and epsilon, respectively, on the 2 node system; $\times20.3$, $\times4.4$, and $\times7.3$, respectively, on the 4 node system.

We observe that the only multi-threaded scikit-learn solver (\texttt{lbfgs}) performs worse than the other scikit-learn solvers, both in test loss and time to converge; \texttt{liblinear} is the best choice.
The only exception to this is higgs on the 2 node system, where \texttt{lbfgs} performs $\times4$ faster than \texttt{liblinear}, but operates at a higher test loss.
H2O performance is somewhat extreme: its multi-threaded solver takes over all the cores in the system and is able to achieve the expected test loss, but the performance varies dramatically across datatasets.
For higgs, it is second only to \texttt{snap.ml MT}, and significantly faster than the alternatives ($\times25$ faster than scikit-learn on the 2 node machine).
However, for epsilon, it is by far the slowest solver (by at least $\times3$).
We expect this to be an issue with large number of features (epsilon has $2k$): by artificially reducing the number of features to 200 using the \texttt{max\_active\_predictors} H2O parameter, we get an order of magnitude speedup in time, with, however, a dramatic degradation of the test loss.

\vspace{-0.1in}
\section{Conclusion}
\vspace{-0.1in}
We have started with a  state-of-the-art  multi-threaded implementation of SDCA. We proposed several modifications to make it more aligned with the hardware architecture of a modern CPU while preserving good convergence behavior. 1) we have proposed a bucketing approach to improve the memory access pattern of a randomized stochastic algorithm, 2) we have introduced a novel dynamic data partitioning scheme to alleviate issues of false share between threads and 3) we have proposed a hierarchical numa-aware partitioning of the workload across threads to enable efficient scaling of the algorithms across numa-nodes.
Combining all these optimizations we achieve a gain of up to $\times12$ compared to a state-of-the-art optimized system-agnostic parallel implementation of the same algorithm.


\bibliographystyle{plain}
\bibliography{ms}

\end{document}